\documentclass{llncs}
\usepackage{subfig}
\usepackage{algorithm}
\usepackage{algorithmic}
\usepackage{hyperref}
\usepackage{amsmath}
\usepackage{amssymb}
\usepackage{graphicx}
\usepackage{enumitem}
\usepackage{wrapfig}

\DeclareMathOperator*{\argmin}{arg\,min}

\begin{document}

\title{Unsupervised Probabilistic Deformation Modeling for Robust Diffeomorphic Registration}

\author{*}
\institute{*}
\author{Julian Krebs\inst{1}\inst{2} \and Tommaso Mansi\inst{2} \and Boris Mailh\'{e}\inst{2} \and Nicholas Ayache\inst{1} \and \\ Herv\'{e} Delingette\inst{1}}


\institute{Inria, Epione Team, Universit\'{e} C\^{o}te d'Azur, Sophia Antipolis, France \and Siemens Healthineers, Medical Imaging Technologies, Princeton, NJ, USA}

\maketitle
\vspace{-6pt}

\begin{abstract} 

We propose a deformable registration algorithm based on unsupervised learning of a low-dimensional probabilistic parameterization of deformations. We model registration in a probabilistic and generative fashion, by applying a conditional variational autoencoder (CVAE) network. This model enables to also generate normal or pathological deformations of any new image based on the probabilistic latent space. Most recent learning-based registration algorithms use supervised labels or deformation models, that miss important properties such as diffeomorphism and sufficiently regular deformation fields. In this work, we constrain transformations to be diffeomorphic by using a differentiable exponentiation layer with a symmetric loss function. We evaluated our method on 330 cardiac MR sequences and demonstrate robust intra-subject registration results comparable to two state-of-the-art methods but with more regular deformation fields compared to a recent learning-based algorithm. Our method reached a mean DICE score of 78.3\% and a mean Hausdorff distance of 7.9mm. In two preliminary experiments, we illustrate the model's abilities to transport pathological deformations to healthy subjects and to cluster five diseases in the unsupervised deformation encoding space with a classification performance of 70\%. 


\end{abstract}

\vspace{-14pt}
\section{Introduction}
\vspace{-3pt}
Deformable registration is an essential task in medical image analysis. It describes the process of finding voxel correspondences in a pair of images \cite{sotiras}. Traditional registration approaches aim to optimize a local similarity metric between deformed and target image, while being regularized by various energies \cite{sotiras}. In order to retrieve important properties such as invertible deformation fields, diffeomorphic registration was introduced. Among other parametrizations, one way to parametrize diffeomorphisms are stationary velocity fields (SVF) \cite{arsigny2006log}. 

In recent years, major drawbacks of these approaches like high computational costs and long execution times have led to an increasing popularity of learning-based algorithms -- notably deep learning (DL). One can classify these algorithms as supervised or unsupervised. Due to the difficulty of finding ground truth voxel correspondences, supervised methods need to rely on predictions from existing algorithms \cite{yang2016fast}, simulations \cite{sokooti2017nonrigid} or both \cite{krebs2017robust}. These methods are either limited by the performance of the used existing algorithms or the realism of simulations. On the other hand, unsupervised approaches make use of spatial transformer layers (STN \cite{jaderberg2015spatial}) to warp the moving image in a differentiable way such that loss functions can operate on the warped image (similarity metric) and on the transformation itself (regularization) \cite{balakrishnan2018unsupervised,jason2016back,de2017end}. While unsupervised approaches perform well in minimizing a similarity metric, it remains unclear if the retrieved deformation fields are sufficiently regular which is of high interest for intra-subject registration. Furthermore, important properties like symmetry or diffeormorphisms \cite{sotiras} are still missing in DL-based approaches.

In this paper, we suggest to learn a low-dimensional probabilistic parameterization of deformations which is restricted to follow a prescribed distribution. This stochastic encoding is defined by a latent code vector of an encoder-decoder neural network and it restricts the space of plausible deformations with respect to the training data. By using a conditional variational autoencoder (CVAE \cite{kingma2014semi}), our generative network constrains encoder and decoder on the moving image. After training, the probabilistic encoding can be potentially used for deformation analysis tasks such as clustering of deformations or the generation of new deformations for a given image -- similar to the deformations seen during training. Furthermore, we include a generic vector field exponentiation layer to generate diffeomorphic transformations. Our framework contains an STN and can be trained with a choice of similarity metrics. To avoid asymmetry, we use a symmetric local cross correlation criterion. The main contributions are:
\begin{itemize}[noitemsep]
\vspace{-2pt}
\item A probabilistic formulation of the registration problem through unsupervised learning of an encoded deformation model.
\item A differentiable exponentiation and an user-adjustable smoothness layer that ensure the outputs of neural networks to be regular and diffeomorphic.
\item As a proof of concept, first experiments on deformation transport and disease clustering.
\vspace{-2pt}
\end{itemize}

\section{Methods}
\vspace{-3pt}
The goal of image registration is to find the spatial transformation $\mathcal{T}_z:\mathbb{R}^3\rightarrow\mathbb{R}^3$, parametrized by a $d$-dimensional vector $z\in\mathbb{R}^d$, which best warps the moving image $\mathbf{M}$ to match the fixed image $\mathbf{F}$. Both images are defined in the spatial domain $\Omega\in\mathbb{R}^3$. Typically, this is done by minimizing an objective function of the form: $\argmin_z  \mathcal{F}(z,\mathbf{M},\mathbf{F})= \mathcal{D}\left(\mathbf{F},\mathbf{M} \circ\ \mathcal{T}_z) + \mathcal{R}(\mathcal{T}_z\right)$ with the image similarity $\mathcal{D}$ of the fixed $\mathbf{F}$ and the warped moving image $\mathbf{M} \circ\ \mathcal{T}_z$ and a spatial regularizer $\mathcal{R}$. Recent unsupervised DL-based approaches (e.g.\ \cite{balakrishnan2018unsupervised,jason2016back}) mimic the optimization of such an objective function. 

\begin{figure}[tb]
\centering 
\subfloat[]{\includegraphics[trim=21 282 515 2,clip,width=0.65\linewidth]{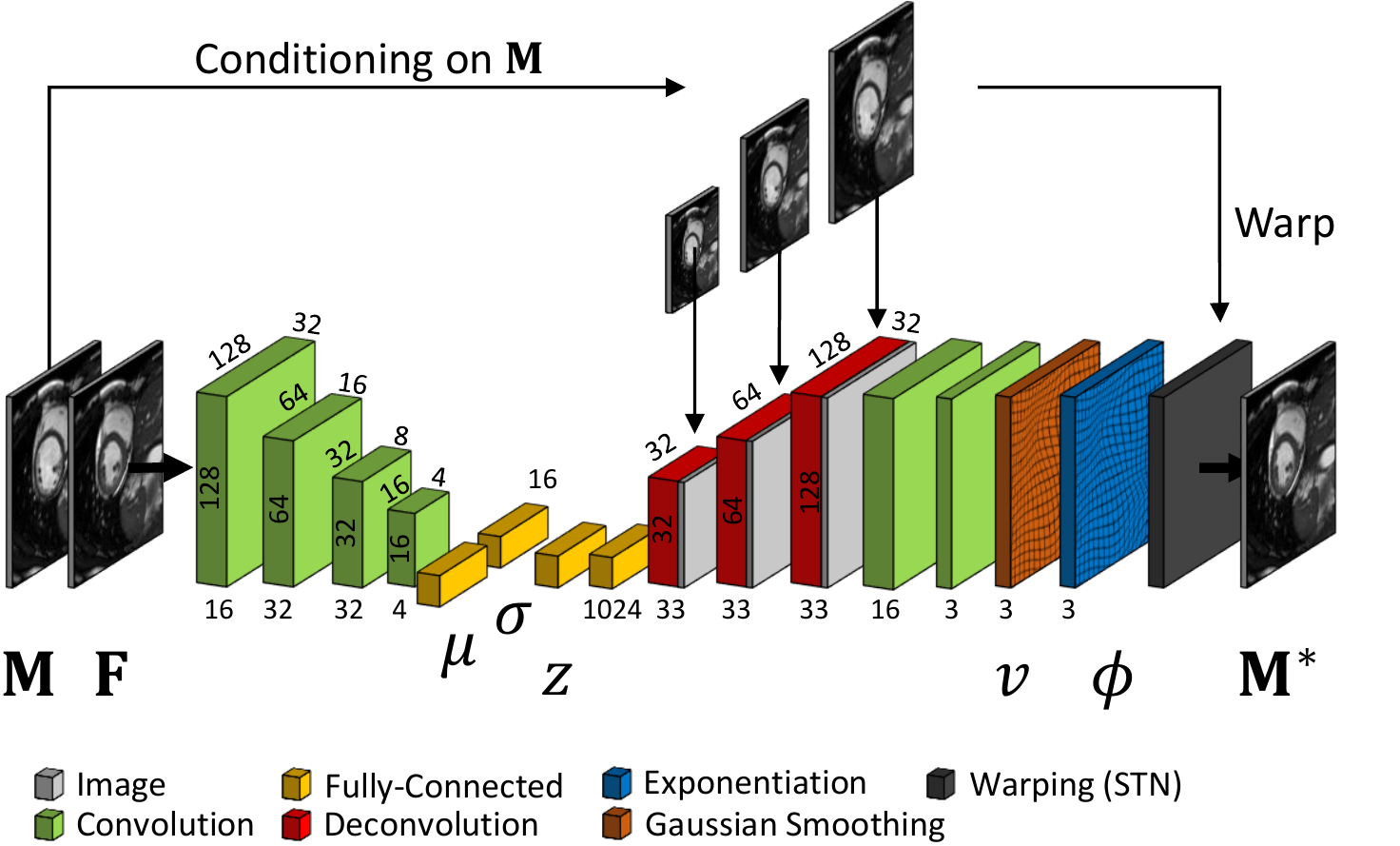}\label{architecture}}\subfloat[]{\includegraphics[trim=160 282 533 2,clip,width=0.35\linewidth]{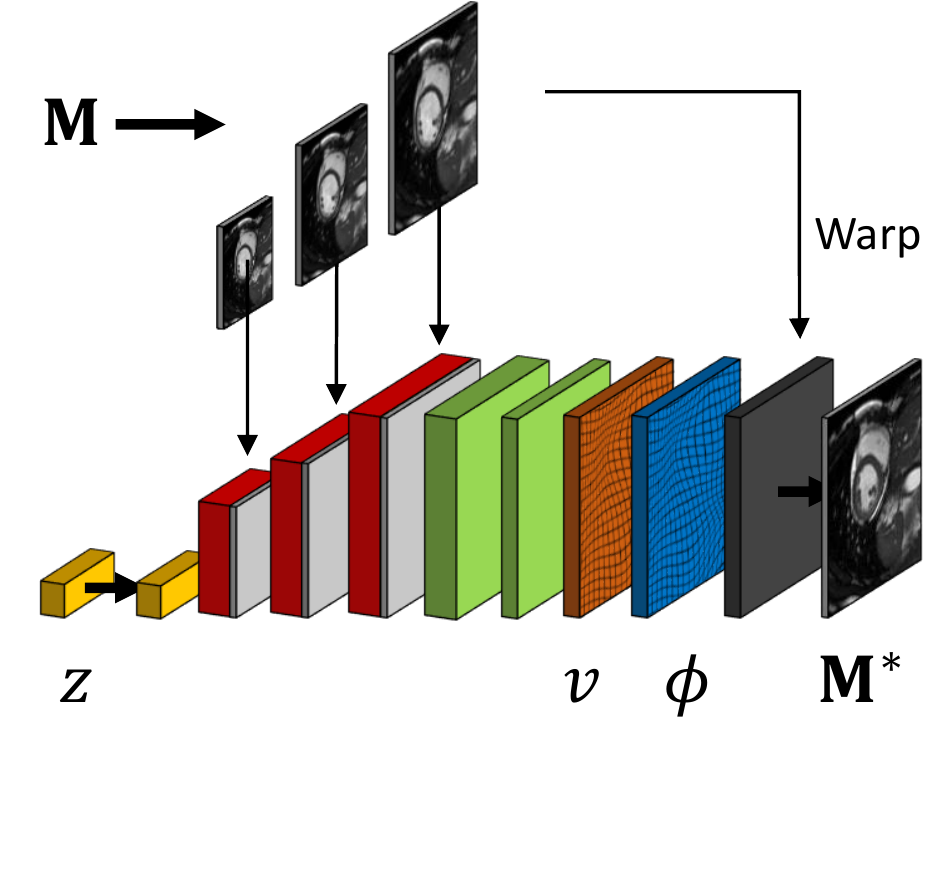}\label{decoder}}
\vspace{-5pt}
\caption{\small{(a) CVAE registration network during training and registration including diffeomorphic layer (exponentiation). Deformations are encoded in $z$ from which velocities are decoded while being conditioned on the moving image. (b) Decoder network for sampling and deformation transport: Apply $z$-code conditioned on any new image $\mathbf{M}$.}}
\vspace{-7pt}
\end{figure}

Instead, we propose to model the registration probabilisitcally by parametrizing the deformation as a vector $z$ to follow a prior $p(z)$. To learn this probabilistic space, we define the latent vector of dimensionality $d$ in an encoder-decoder neural network as this $z$. Given the moving and the fixed image as input, a variational inference method (CVAE \cite{kingma2014semi}) is used to \emph{reconstruct} the fixed by warping the moving image. An exponentiation layer interprets the network's output as velocities $v$ (an SVF) and returns a diffeomorphism $\phi$ which is used by a dense STN to retrieve the warped image $\mathbf{M}^*$. To enforce an user-adjustable level of deformation smoothness (comparable to \cite{lorenzi2013lcc}), a convolutional Gaussian layer is added before the exponentiation with Gaussian weights according to the variance $\sigma^2_S$. During training, the network parameters are updated through back-propagation of the gradients. The network architecture can be seen in Fig.\ \ref{architecture}. Finally, registration is done in a single forward path. The trained probabilistic framework can be also used for the sampling of deformations as shown in Fig. \ref{decoder}. 

\vspace{-4pt}
\subsubsection{Learning a Probabilistic Deformation Encoding}
Learning a generative model typically involves a latent variable model (as in VAE), where an encoder maps an image to its $z$-code -- a low-dimensional latent vector, from which a decoder aims to reconstruct the original image. Typically, the encoder and decoder are defined as distributions $q_\omega$ and $p_\gamma$ with trainable network parameters $\omega$ and $\gamma$. The network is trained by maximizing a lower bound on the data likelihood with respect to a prior distribution $p(z)$. We define the prior as multivariate unit Gaussians $p(z)=\mathcal{N}(0,I)$ with the identity matrix $I$. In CVAE \cite{kingma2014semi}, encoder $q_\omega$ and decoder $p_\gamma$ distributions are additionally conditioned on extra information (e.g.\ classes). We propose to frame image registration as a reconstruction problem in which the moving image $\mathbf{M}$ acts as the conditioning data and is warped to reconstruct or to match the fixed image $\mathbf{F}$. Thus, the decoder reconstructs $\mathbf{F}$ given $z$ and $\mathbf{M}$: $p_\gamma(\mathbf{F}\mid z, \mathbf{M})$. To have $z$, the encoder serves as an approximation of the intractable true posterior probability of $z$ given $\mathbf{F}$ and $\mathbf{M}$ and is denoted as $q_\omega(z\mid \mathbf{F},\mathbf{M})$. Since the prior $p(z)$ is defined as multivariate unit Gaussians, the encoder network predicts the mean $\mu \in \mathbb{R}^d$ and diagonal covariance $\sigma \in \mathbb{R}^d$, from which $z$ is drawn: $q_\omega(z\mid \mathbf{F},\mathbf{M})=\mathcal{N}(\mu(\mathbf{F},\mathbf{M}),\sigma(\mathbf{F},\mathbf{M}))$. 

Both distributions can be combined in a two-term loss function \cite{kingma2014semi} where the first term describes the reconstruction loss as the expected negative log-likelihood of $p_\gamma(\mathbf{F}\mid z, \mathbf{M})$. In other words, the reconstruction loss represents a similarity metric between input $\mathbf{F}$ and output $\mathbf{M}^*$. The second term acts as a regularization term on the deformation latent space by forcing the encoded distribution $q_\omega(z\mid \mathbf{F},\mathbf{M})$ to be close to the prior probability distribution $p(z)$ using a Kullback-Leibler (KL) divergence. The loss function results in:
\begin{equation}
l(\omega, \gamma, \mathbf{F}, \mathbf{M}) = -E_{z\sim q_\omega(\cdot|\mathbf{F}, \mathbf{M})}\left[\text{log} p_\gamma(\mathbf{F}\mid z, \mathbf{M})\right] + KL\left[q_\omega(z\mid \mathbf{F},\mathbf{M})\parallel p(z)\right],\vspace{-3pt}
\end{equation}
where the KL-divergence can be computed in closed form \cite{kingma2014semi}. Assuming a Gaussian log-likelihood term of $p_\gamma$ is equivalent to minimizing a weighted SSD criterion (cf.\ \cite{kingma2014semi}). We propose instead to use a symmetric local cross-correlation (LCC) criterion due to its favorable properties for registration \cite{lorenzi2013lcc} and assume a LCC Boltzmann distribution $p_\gamma(\mathbf{F}\mid z, \mathbf{M}) \sim \exp(-\lambda\mathcal{D}_\textit{LCC}(\mathbf{F},\mathbf{M},v))$ with the LCC criterion $\mathcal{D}_\textit{LCC}$ and the weighting factor $\lambda$. Using the velocities $v$ and a small constant $\epsilon$, which is added for numerical stability, we define:
\begin{equation}
\mathcal{D}_\textit{LCC}(\mathbf{F},\mathbf{M},v) = \frac{1}{P} \sum_{x \in \Omega}  \frac{\overline{\mathbf{F}_x\circ \text{exp}\left(-\frac{v_x}{2}\right)\mathbf{M}_x\circ \text{exp}\left(\frac{v_x}{2}\right)}^2}{\overline{\left[\mathbf{F}_x\circ \text{exp}\left(-\frac{v_x}{2}\right)\right]^2}\hspace{1mm} \overline{\left[\mathbf{M}_x\circ \text{exp}\left(\frac{v_x}{2}\right)\right]^2} + \epsilon},
\end{equation}
with a total number of $P$ pixels $x\in \Omega$ and where $\bar{\cdot}$ symbolizes the local mean image derived by Gaussian smoothing with a strength of $\sigma_G$ and kernel size $k$. To help the reconstruction task, we introduce conditioning by involving $\mathbf{M}$ not only as the image to be warped in the STN, but also in the first decoder layers by concatenating down-sampled versions of $\mathbf{M}$ with the filter maps on each scale. The hypothesis is that in order to better optimize the reconstruction loss, the network makes use of the provided extra information of $\mathbf{M}$ such that less anatomical but more deformation information are conveyed by the low-dimensional latent layer, which would make the encoding more geometry-invariant.
\vspace{-4pt}

\subsubsection{Exponentiation Layer: Generating Diffeomorphisms} In the SVF setting, the transformation $\phi$ is defined as the Lie group exponential map with respect to the velocities $v$: $\phi(x) = \text{exp}(v)$. For efficient computation, the scaling and squaring algorithm is typically used \cite{arsigny2006log}. In order to generate diffeomorphic transformations $\phi$ in a neural network, we propose an exponentiation layer that implements this algorithm in a fully differentiable way. To this end, the layer expects a vector field as input (the velocities $v$) which is scaled with a factor $N$ which we precompute on a subset of the training data according to the formulations in \cite{arsigny2006log}. In the squaring step, the approximated $\phi_0\approx id + v*2^{-N}$ (with $id$ as a regular grid) is recursively squared, $N$-times, from $k=1$ to $N$: $\phi_k = \phi_{k-1} \circ \phi_{k-1}$. The result is the diffeomorphism $\phi_N\equiv\phi$ \cite{arsigny2006log}. The squaring step requires the composition of two vector fields on regular grids which we realized by linear interpolation. All these computations consist of standard operations that can be added to the computational graph and are auto-differentiable in modern deep learning libraries. This differentiable layer can be added to any neural network which predicts (stationary) velocity fields.
\vspace{-1pt}
\section{Experiments}
\vspace{-1pt}
We evaluate our framework on an intra-subject task of cardiac MRI cine registration where end-diastole frames are registered to end-systole frames (ED-ES) -- a very large deformation. Furthermore, we show preliminary experiments evaluating the learned deformation encoding: its potentials for transporting encoded deformations from one subject to another and showing the clustering of diseases in the encoding space. All experiments are in 3-D.
\begin{figure}[tb]
\centering 
\includegraphics[trim=9 337 258 8,clip,width=\linewidth]{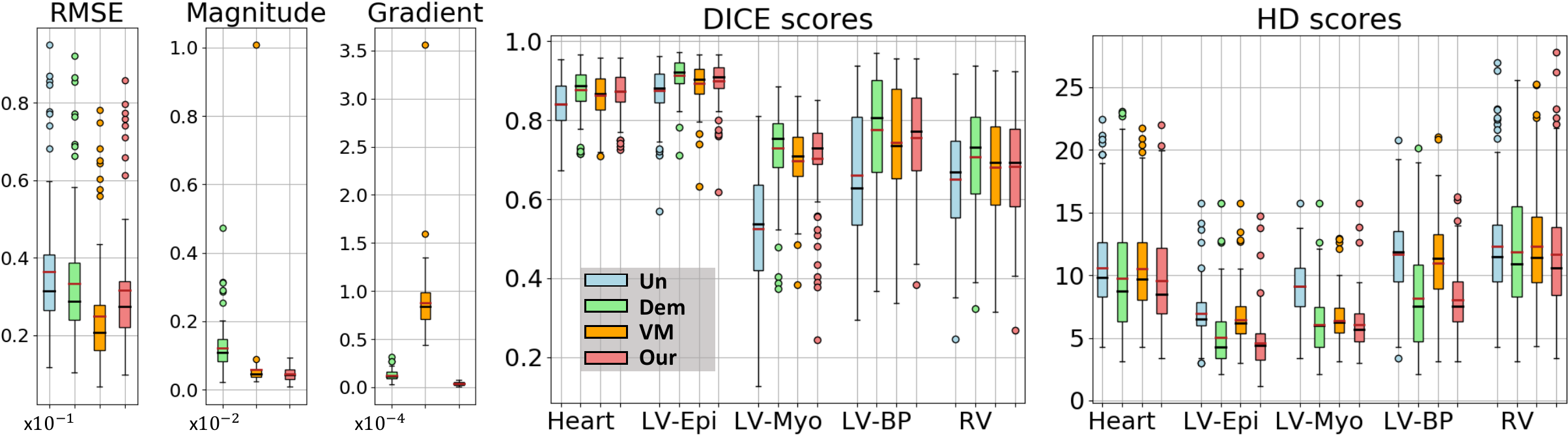}
\\
\vspace{-5pt}
\caption{\small{Comparing registration performance: unregistered (Un), LCC-Demons (Dem), VoxelMorph (VM) and our method in terms of RMSE and mean deformation magnitude and gradient, DICE and 95\%-tile Hausdorff distances (HD).}}\label{resultscardiac}
\vspace{-8pt}
\end{figure}

We used 184 short-axis datasets acquired from different hospitals and 150 cases from the Automatic Cardiac Diagnosis Challenge (ACDC) at STACOM 2017\footnote{https://www.creatis.insa-lyon.fr/Challenge/acdc/index.html}, mixing congenital heart diseases with images from adults. We used 234 cases for training and for testing the remaining 100 cases from ACDC, that contain segmentation and disease label information from five cardiac diseases. Both information were only used for evaluation purposes. All images were sampled with a spacing of $1.5 \times 1.5 \times 3.15$ mm and cropped to a size of $128\times128\times32$ voxels. These dimensions were chosen to save computation time and are not a limitation of the framework (validated on different image sizes). 
\vspace{-4pt}

\subsubsection{Implementation Details} The encoder of our neural network consisted of four convolutional layers with strides (2, 2, 2, 1) (Fig.\ \ref{architecture}). The bottleneck layers ($\mu$, $\sigma$, $z$) were fully-connected. The decoder had one fully-connected and three deconvolutional layers, where the outputs at each layer were concatenated with sub-sampled versions of $\mathbf{M}$. Two convolutional layers and a convolutional Gaussian layer with $\sigma_S=3$ (kernel size 15) were placed in front of the exponentiation and transformer layer. The latent code size $d$ was set to 16 as a trade off between registration quality and generalizability. This leads to a total of $\sim$267k trainable parameters. L2 weight decay with a factor of 0.0001 was applied. The numbers of iterations in the exponentiation layer was set to $N=4$ in all experiments. In training, the strength of the Gaussians for computing the LCC was set to $\sigma_G=2$ with a kernel size $k=9$. The loss balancing factor $\lambda=5000$ was empirically chosen such that encoded training samples roughly had zero means and variances of 1 and the reconstruction loss was optimized. We used the Adam optimizer with a learning rate of 0.0005 and a batch size of one. We performed online data augmentation by randomly shifting, rotating, scaling and mirroring training images. The framework has been implemented using \textit{Keras} with \textit{Tensorflow}. Training took 24 hours on a \textit{NVIDIA GTX TITAN X} GPU.

\begin{figure}[tb]
\centering 
\includegraphics[trim=1 218 446 1,clip,width=1.0\linewidth]{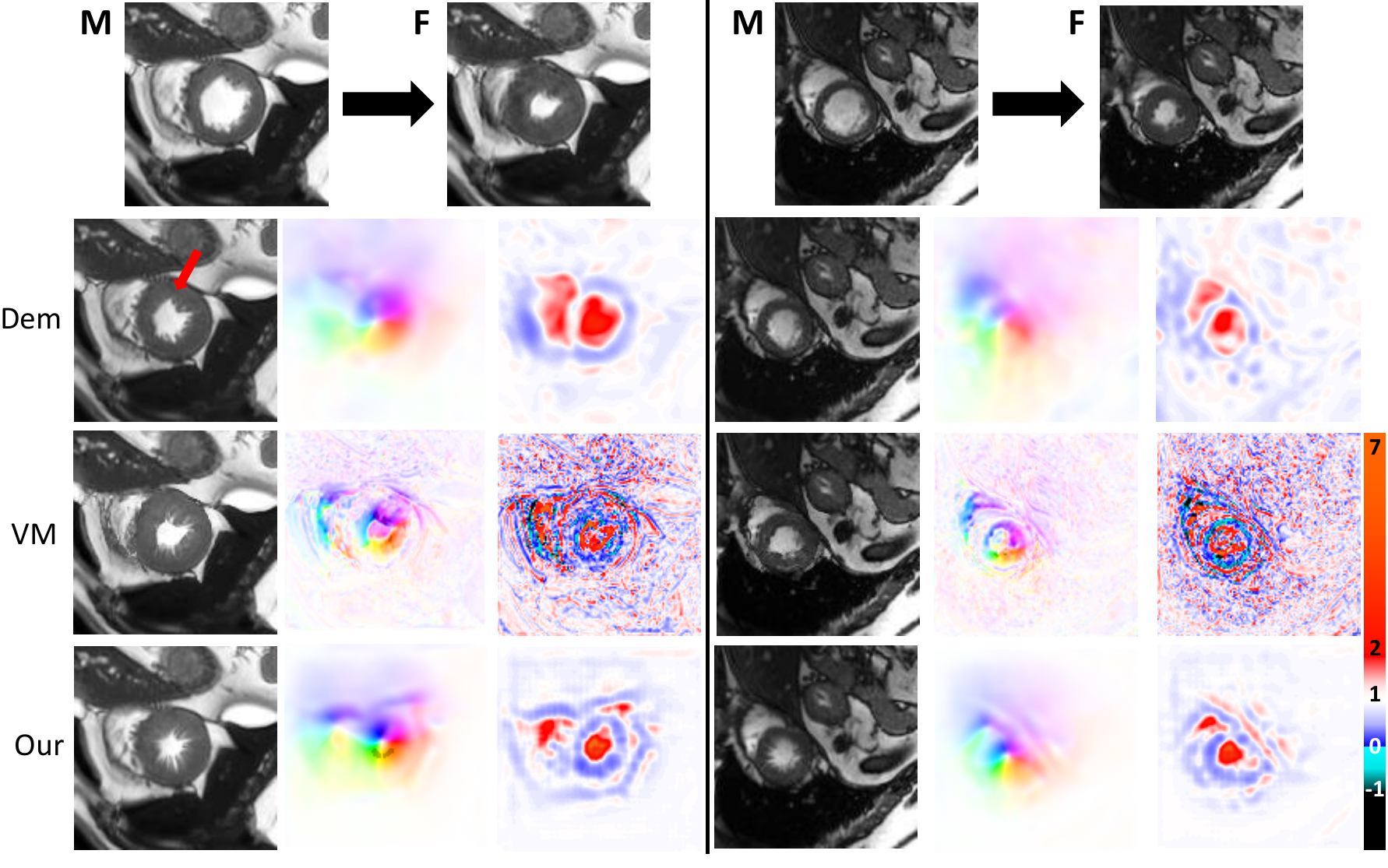}
\vspace{-18pt}
\caption{\small{Two random examples of end-diastole to end-systole registration: (Row 1) original images. The LCC-demons (Dem, Row 2) and VoxelMorph (VM, Row 3) versus our method (Row 4), showing the warped moving image, the deformation field and the Jacobian determinants. All results are in 3-D, showing the central short-axis slices.}}\label{resultimages}
\vspace{-13pt}
\end{figure}

\subsubsection{Registration Results}
We compare our registration algorithm with the LCC-demons \cite{lorenzi2013lcc} with manually tuned parameters (on training images) and the non-diffeomorphic DL-based method VoxelMorph-2 \cite{balakrishnan2018unsupervised} (VM) with a regularization weighting parameter of 1.5, as recommended. As a surrogate measure of registration performance, we used the intensity root mean square error (RMSE), mean DICE score and 95\%-tile Hausdorff distance (HD) in mm on the following anatomical structures: myocardium and epicardium of the left ventricle  (LV-Epi, LV-Myo), left bloodpool (BP) and heart (Heart). The LCC-demons showed better mean DICE scores (averaged over the five structures, in \%) with 79.9 compared to our algorithm with 78.3 and VM with 77.5 (cf.~Fig.~\ref{resultscardiac}). The VoxelMorph algorithm reached a very low RMSE of 0.025 compared to ours (0.031) and the demons (0.034), but could not reach the other algorithms in terms of HD with a mean score of 9.4mm compared to ours with 7.9mm and the demons with 8.2mm. Besides these metrics, VM produced very irregular and highly non-diffeomorphic deformation fields since 2.2\% of the displacements had a negative Jacobian determinant (cf.\ in Fig.\ \ref{resultimages}). In general, our approach led to deformation fields with both smaller amplitudes and smaller gradients than the demons and the VM algorithm. Furthermore, our results were more robust as variances were lower for all metrics compared to the demons and lower or comparable to VM. This is also visible in Fig.\ \ref{resultimages} and further shown by the fact that HD scores are the smallest experienced in the experiments. Average execution time per test case was 0.32s using the mentioned GPU and an \emph{Intel Xeon CPU E5}, compared to 108s for the demons on CPU.
\begin{figure}[t]
\centering 
\includegraphics[trim=1 238 421 17,clip,width=1.\linewidth]{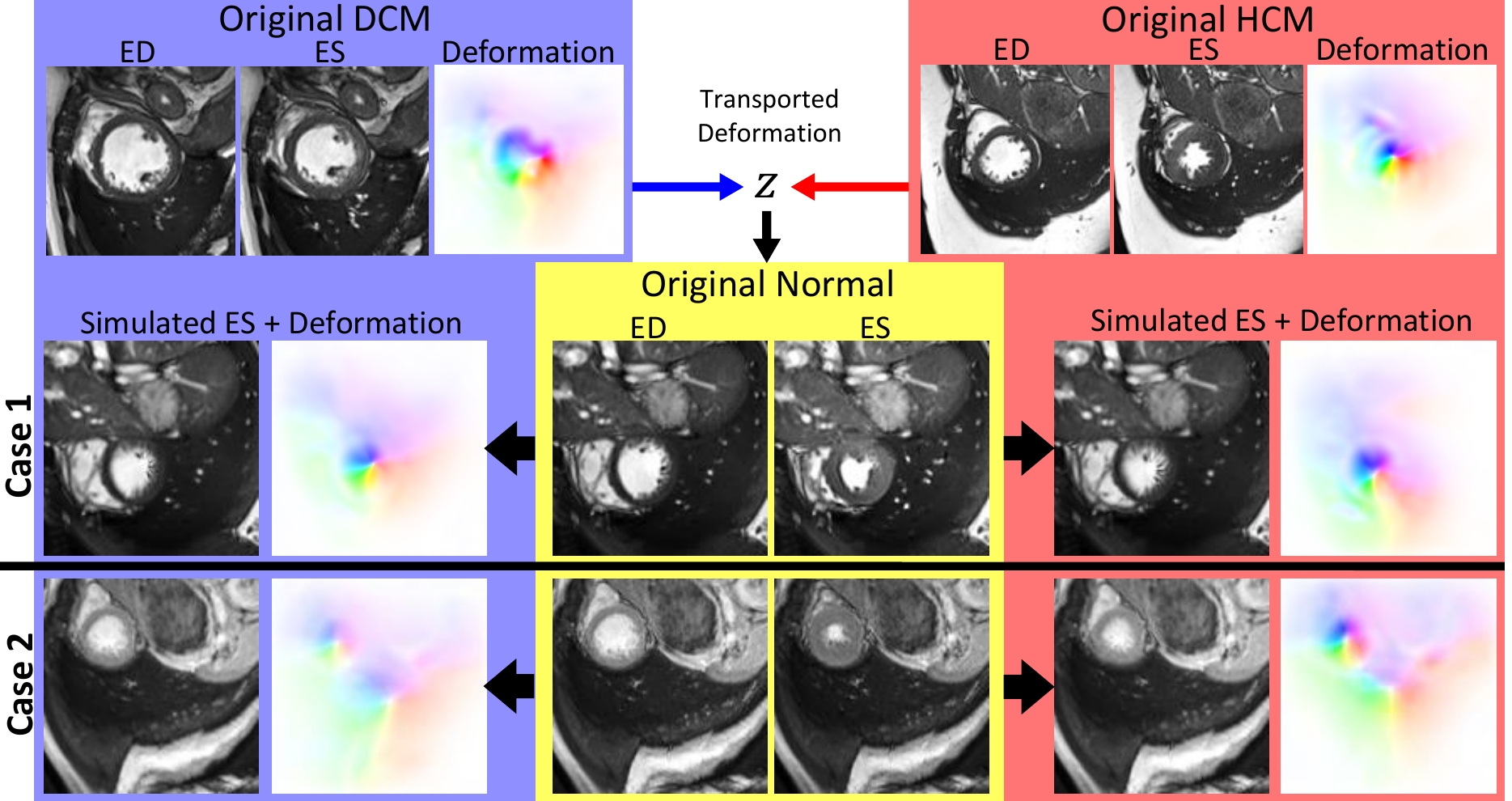}
\vspace{-10pt}
\caption{\small{Transport the $z$-code of pathological deformations (top row: cardiomyopathy DCM and hypertrophy HCM) to two healthy subjects (bottom rows: Normal). The simulated deformation fields are \emph{similar} compared to the pathological deformations but are adapted to the geometry of the healthy image (e.g.\ translated).}}\label{transport}
\vspace{-6pt}
\end{figure}

\subsubsection{Deformation Encoding} 
\begin{wrapfigure}[12]{r}{0.4\textwidth}
\centering 
\vspace{-2pt} 
\includegraphics[trim=38 13 45 40,clip,width=0.35\textwidth]{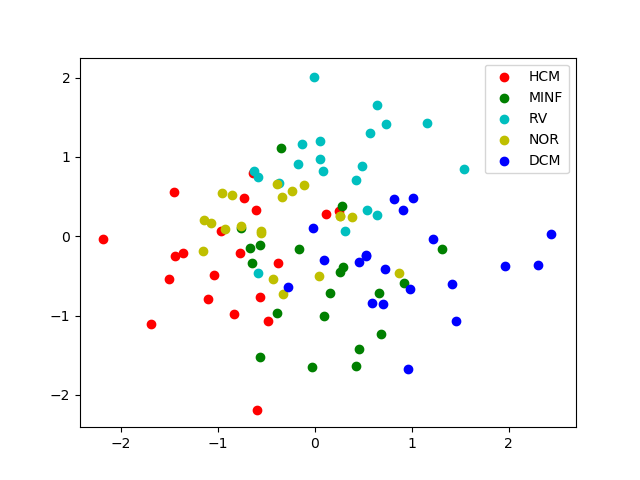}
\vspace{-10pt}
\caption{\small{Distribution of cardiac diseases after projecting 100 $z$-codes of test images on 2 CCA components.}}\label{ldaplot}
\end{wrapfigure} 
For evaluating the learned deformation encoding, we show geometry-invariance by transporting a deformation from one subject to another. Therefore, we take a $z$-code from a pathological subject and condition the decoder on the ED image of healthy subjects (Fig.\ \ref{decoder}). More precisely, in Fig.\ \ref{transport} we transported a cardiomyopathy (DCM) and hypertrophy (HCM) deformation to two healthy cases (Normal). One can see the disease-specific deformation (DCM: reduced cardiac contraction) which are different from the healthy transformations. The resulting deformation fields are adapted to the anatomy of the conditioning image and they are translation-invariant.

In a second experiment, we used the encoded $z$-codes and disease information of our cardiac test set to visualize the structure of the learned space. Therefore, we linearly projected the 16-D $z$-codes to a 2-D space by using the two most discriminative CCA components (canonical correlation analysis). We used the ACDC classes: dilated cardiomyopathy DCM, hypertrophic cardiomyopathy HCM, myocardial infarction MNF, abnormal right ventricle RV and normal NOR.  In Fig.\ \ref{ldaplot}, one can see that the classes of the 100 test sets are clustered in the projected space. The five class classification accuracy reaches 70\% with 10-fold cross-validation, by using the six most discriminative CCA components and applying support vector machine (SVM) on-top. These results which are solely based on unsupervised deformation $z$-codes suggest that similar deformations are close to each other in the deformation encoding space.
\vspace{-4pt}

\section{Conclusion}
\vspace{-4pt}
We presented an unsupervised deformable registration approach that learns a probabilistic deformation encoding. This encoding constrains the registration and leads to robust and accurate registration results on a large dataset of cardiac images. Furthermore, an exponentiation layer has been introduced that creates diffeomorphic transformations. The performance of the proposed method was comparable and partially superior to two state-of-the-art algorithms. Our approach produced more regular deformation fields than a DL-based algorithm. Furthermore, first results show, that the probabilistic encoding could potentially be used for deformation transport and clustering tasks. In future work, we plan to further explore the deformation encoding to evaluate these tasks more deeply.

\vspace{6pt}
\noindent \textbf{Acknowledgements: } Data used in this article were obtained from the EU FP7-funded project MD-Paedigree and the ACDC STACOM challenge 2017.
  
\noindent \textbf{Disclaimer: } This feature is based on research, and is not commercially available. Due to regulatory reasons its future availability cannot be guaranteed.
\vspace{-5pt}
\bibliography{32_dlmia18.bib}

\begin{thebibliography}{10}
\providecommand{\url}[1]{\texttt{#1}}
\providecommand{\urlprefix}{URL }

\bibitem{arsigny2006log}
Arsigny, V., Commowick, O., Pennec, X., Ayache, N.: A log-euclidean framework
  for statistics on diffeomorphisms. In: International Conference on Medical
  Image Computing and Computer-Assisted Intervention. pp. 924--931. Springer
  (2006)

\bibitem{balakrishnan2018unsupervised}
Balakrishnan, G., et~al.: An unsupervised learning model for deformable medical
  image registration. In: Proceedings of the IEEE CVPR. pp. 9252--9260 (2018)

\bibitem{jaderberg2015spatial}
Jaderberg, M., Simonyan, K., Zisserman, A., et~al.: Spatial transformer
  networks. In: Advances in neural information processing systems. pp.
  2017--2025 (2015)

\bibitem{jason2016back}
Jason, J.Y., et~al.: Back to basics: Unsupervised learning of optical flow via
  brightness constancy and motion smoothness. In: ECCV. pp. 3--10. Springer
  (2016)

\bibitem{kingma2014semi}
Kingma, D.P., et~al.: Semi-supervised learning with deep generative models. In:
  Advances in Neural Information Processing Systems. pp. 3581--3589 (2014)

\bibitem{krebs2017robust}
Krebs, J., Mansi, T., Delingette, H., et~al.: Robust non-rigid registration
  through agent-based action learning. In: International Conference on Medical
  Image Computing and Computer-Assisted Intervention. pp. 344--352. Springer
  (2017)

\bibitem{lorenzi2013lcc}
Lorenzi, M., Ayache, N., Frisoni, G.B., et~al.: {LCC-D}emons: a robust and
  accurate symmetric diffeomorphic registration algorithm. NeuroImage  81,
  470--483 (2013)

\bibitem{sokooti2017nonrigid}
Sokooti, H., de~Vos, B., et~al.: Nonrigid image registration using multi-scale
  3d convolutional neural networks. In: International Conference on Medical
  Image Computing and Computer-Assisted Intervention. pp. 232--239. Springer
  (2017)

\bibitem{sotiras}
Sotiras, A., Davatzikos, C., Paragios, N.: Deformable medical image
  registration: A survey. IEEE Transactions on Medical Imaging  32(7),
  1153--1190 (2013)

\bibitem{de2017end}
de~Vos, B.D., Berendsen, F.F., Viergever, M.A., Staring, M., I{\v{s}}gum, I.:
  End-to-end unsupervised deformable image registration with a convolutional
  neural network. In: Deep Learning in Medical Image Analysis and Multimodal
  Learning for Clinical Decision Support, pp. 204--212. Springer (2017)

\bibitem{yang2016fast}
Yang, X., Kwitt, R., Niethammer, M.: Fast predictive image registration. In:
  Deep Learning and Data Labeling for Medical Applications, pp. 48--57.
  Springer (2016)

\end{thebibliography}

\end{document}